\documentclass{scrartcl}

\usepackage[top=3cm, bottom=3.5cm, left=3cm, right=3cm]{geometry}
\usepackage{graphicx} 
\usepackage{authblk}

\RequirePackage{graphicx}
\PassOptionsToPackage{hyphens}{url}\RequirePackage{hyperref}
\RequirePackage{lmodern}
\usepackage[utf8]{inputenc} 

\usepackage[acronym]{glossaries}
\usepackage{xcolor}
\usepackage{arydshln}
\usepackage{multirow}
\usepackage{adjustbox}

\title{Transformer Training Strategies for Forecasting Multiple Load Time Series}
\author[1*]{Matthias~Hertel}
\author[1]{Maximilian~Beichter}
\author{Benedikt~Heidrich}
\author{Oliver~Neumann}
\author{Benjamin~Schäfer}
\author{Ralf~Mikut}
\author{Veit~Hagenmeyer}
\affil[1]{Institute for Automation and Applied Informatics\protect\\ Karlsruhe Institute of Technology}
\affil[*]{matthias.hertel@kit.edu}
\date{}

\definecolor{kitpurple}{rgb}{0.64, 0.06, 0.49}

\newcommand{\mv}{multivariate}
\newcommand{\Mv}{Multivariate}
\newcommand{\loc}{local}
\newcommand{\Loc}{Local}
\newcommand{\glob}{global}
\newcommand{\Glob}{Global}

\newcommand{\citelongterm}{\cite{informer, autoformer, fedformer, patchtst, transformers-effective}}

\newcommand{\diff}[1]{#1}
\newcommand{\difffigure}[1]{#1}

\begin{document}

\maketitle

\begin{abstract} 
  In the smart grid of the future, accurate load forecasts on the level of individual clients can help to balance supply and demand locally and to prevent grid outages.
  While the number of monitored clients will increase with the ongoing smart meter rollout, the amount of data per client will always be limited.
  We evaluate whether a Transformer load forecasting model benefits from a transfer learning strategy, where a \glob\ univariate model is trained on the load time series from multiple clients.
  In experiments with two datasets containing load time series from several hundred clients, we find that the \glob\ training strategy is superior to the \mv\ and \loc\ training strategies used in related work. On average, the global training strategy results in 21.8\% and 12.8\% lower forecasting errors than the two other strategies, measured across forecasting horizons from one day to one month into the future.
  A comparison to linear models, multi-layer perceptrons and LSTMs shows that Transformers are effective for load forecasting when they are trained with the \glob\ training strategy.
\end{abstract}

\section*{Introduction}

Climate change is one of the biggest challenges facing humanity, with the risk of dramatic consequences if certain limits of warming are exceeded \cite{ipcc}. To mitigate climate change, the energy system must be decarbonized. A difficulty in decarbonization is that renewable energy supply fluctuates depending on the weather. However, supply and demand must be balanced in the grid at every moment to prevent outages \cite{machowski-bialek}. In addition, with the ongoing decentralization of the renewable energy supply and the installation of large consumers, such as electric vehicle chargers and heat pumps, low-voltage grids are expected to reach their limits \cite{all-electrical-society}.
\diff{
Thus, to balance the grid and to avoid congestions, advanced operation and control mechanisms must be installed in the smart grid of the future \cite{smarts-into-smart-grid, haben-review-low-voltage-load-forecasting}.
This requires accurate forecasts on various aggregation levels, up to fine-grained low-voltage level load forecasts \cite{haben-review-low-voltage-load-forecasting, ordiano}.
Such fine-grained load forecasts can be used for demand-side management, energy management systems, distribution grid state estimation, grid management, storage optimization, peer-to-peer trading, peak shaving, smart electrical vehicle charging, dispatchable feeders, provision of feedback to customers, anomaly detection and intervention evaluation \cite{haben-review-low-voltage-load-forecasting, recent-advances-smart-meter-data, global-cnn, global-models, dispatchable-feeder}. Moreover, the aggregation of fine-grained load forecasts can result in a more accurate forecast of the aggregated load \cite{hong-literature-review-outlook}.

With the smart meter rollout, fine-grained electrical load data will become available for an increasing number of clients. 
In such a scenario where load time series from multiple clients are available, different model training strategies are possible.
The goal of our work is to compare training strategies for the Transformer \cite{transformer}, which was recently used for load forecasting \cite{time-augmented-transformer, ci-paper, tccml-paper, transformer-state-space-model, tft-grid-hierarchies, short-term-load-tft}.
}

\begin{figure}[tb]
    \centering
    \includegraphics[width=0.95\textwidth]{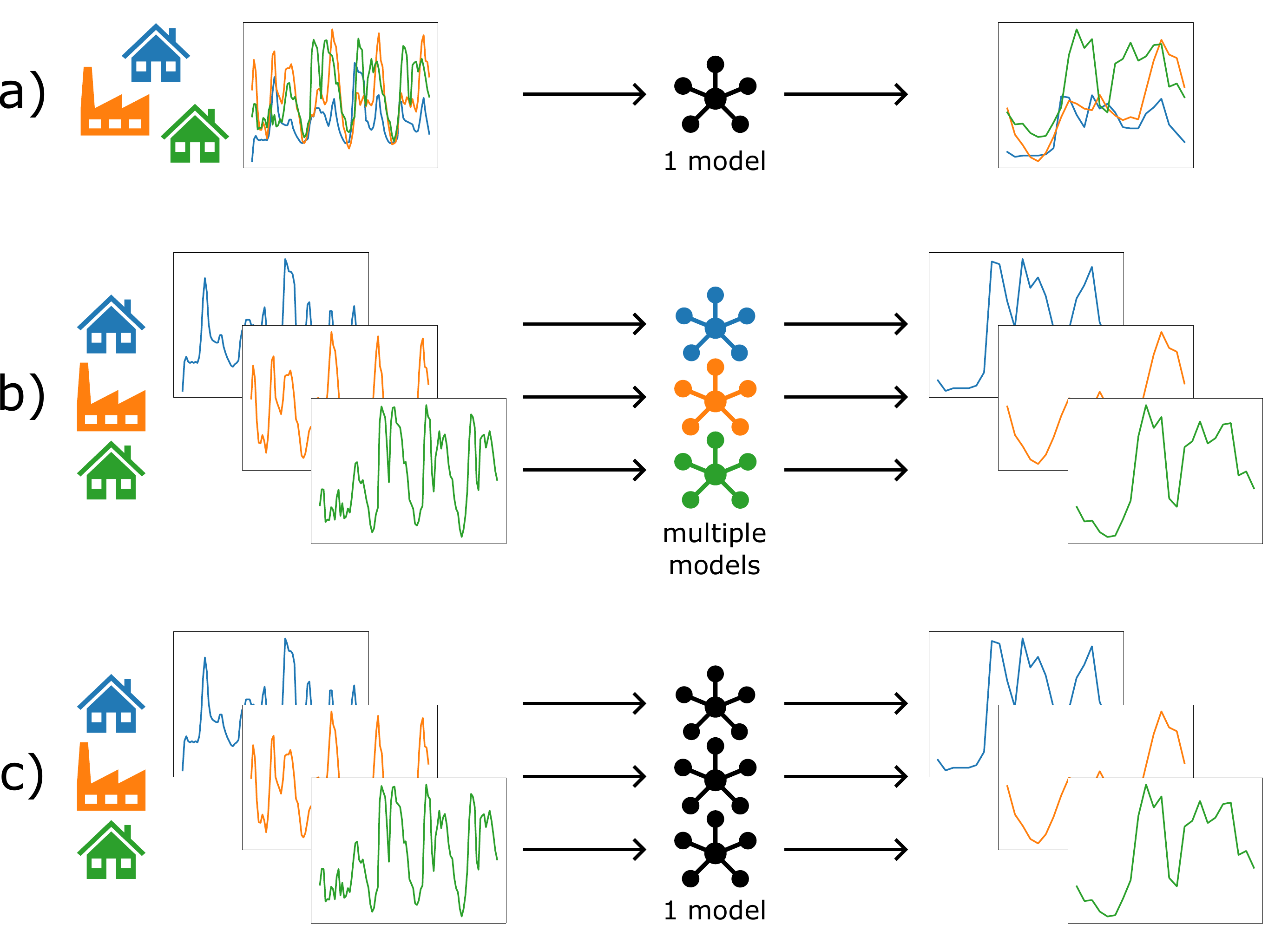}
    \caption{The three training strategies, with models depicted as networks. An example with three load time series, four days input and one day output is shown. (a) \Mv: one model processes all load time series simultaneously; (b) \loc: separate models (blue, orange, green) process each load time series; (c) \glob: one model (black) processes all load time series one at a time.}
    \label{fig:training-strategies}
\end{figure}

\subsection*{Task definition}

We address the following multiple load time series forecasting problem:
At a time step $t$, given the history of the electrical load of $C$ clients $x_0^c, ... x_t^c$ with $1 \leq c \leq C$, the goal is to predict the next $h$ electrical load values $x_{t+1}^c, ..., x_{t+h}^c$ for all clients $1 \leq c \leq C$, where $h$ is called the forecast horizon.

\subsection*{Contribution}

We compare three training strategies for the Transformer in a scenario with multiple load time series. The training strategies are depicted in Figure \ref{fig:training-strategies}. 
\begin{enumerate}
    \item A \emph{\mv}
    model training strategy, where a single model gets all load time series as input and forecasts all load time series simultaneously.
    \item A \emph{\loc} model training strategy, where a separate univariate\footnote{By 'univariate' we mean models which produce a forecast for a single time series. We still call models 'univariate' when they have multiple input variables, such as exogenous time and calendar features.} model is trained for each load time series.
    \item A \emph{\glob} model training strategy, where a generalized univariate model is used to forecast each load time series separately.
\end{enumerate}


\noindent
We compare our models with the models from related work \cite{informer, autoformer, fedformer, patchtst}, as well as with multiple baselines.
In particular, we compare with the linear models used in \cite{transformers-effective}, to figure out if Transformers are effective for load forecasting and which training strategy is the most promising one.

\subsection*{Paper structure}

First, we describe the \nameref{sec:related-work}. Then, the Transformer architecture and the training strategies are described in the \nameref{sec:approach}.
This is followed by the \nameref{sec:experiments}, \nameref{sec:results} and a \nameref{sec:discussion}.
Finally, the paper concludes with the \nameref{sec:conclusion}.

\section*{Related work} \label{sec:related-work}

\diff{
This section first presents related work on long time series forecasting and load forecasting with Transformers.
Most of the load forecasting literature uses \loc\ models, but few works use \glob\ models, which are presented next.
The \glob\ training strategy can be understood as a transfer learning technique. We therefore discuss transfer learning in the field of load forecasting at the end of this section.
}


As Transformer are often used for long time series forecasting with up to one month horizon, various extensions to the Transformer architecture exist that aim to reduce the time and space complexity. This is done by the Informer using ProbSparse self-attention \cite{informer}, by the Autoformer using auto-correlation \cite{autoformer}, by the FEDformer using frequency enhanced decomposition \cite{fedformer} and by PatchTST using patching \cite{patchtst}.
The proposed models are \mv\ or \loc, except for the \glob\ PatchTST \cite{patchtst}.
\diff{The experiments in these works are conducted on six datasets from different domains, including one load forecasting dataset, which we also use in our experiments (see section \nameref{sec:dataset}).}
A \glob\ linear model called LTSF-Linear \cite{transformers-effective} gives better results than the aforementioned \mv\ Transformers.
\diff{Parallel to our work, \glob\ Transformers were shown to beat the aforementioned \mv\ Transformers \cite{iclr-global-transformer}. However, this work does not optimize the model's lookback size and therefore achieves sub-optimal results.}
PatchTST \cite{patchtst} is a \glob\ Transformer with patched inputs and is superior to LTSF-Linear \cite{transformers-effective} on the six datasets.


Transformer architectures for short-term load forecasting are designed to use external calendar and weather features \cite{multienergy, temporal-fusion-transformer}. An evaluation of different architectures is undertaken in \cite{ci-paper}. Further work modifies the architecture for multi-energy load forecasting \cite{multienergy}.
Upstream decompositions are used to improve the forecast quality \cite{ceemdan2023}.
\diff{These models are not compared on a common benchmark dataset, but evaluated on different datasets on city or national level. There, usually only one load time series is available, which only allows for \loc\ models. Furthermore, the models are not compared to the Transformer architectures for long time series.}



Global load forecasting models are already used with convolutional neural networks \cite{global-cnn} and N-BEATS \cite{global-models}.
A mixture between a \mv\ and a \glob\ model is investigated in \cite{pooling-rnn}, where a single recurrent neural network (RNN) model is trained on randomly pooled subsets of the time series.
Some works cluster the time series and then train \glob\ or \mv\ models for each cluster \cite{k-means-lstm, individual-load-forecasting}.
PatchTST \cite{patchtst} is a \glob\ Transformer with patched inputs. We compare to this approach in our experiments.


The authors of \cite{transfer-learning-smart-buildings-review} and \cite{roles_of_transfer_learning} present current literature on transfer learning in the domain of energy systems. They define a taxonomy of transfer learning methods and discuss different strategies of using transfer learning with buildings from different domains. 
Two works \cite{nawar2023transfer, tgdlf2} use transfer learning by pre-training and fine-tuning Transformers.
Transferability from one building to another is tested in \cite{nawar2023transfer}, and from one district to another in \cite{tgdlf2}.
In contrast to these works, our transfer learning approach is to train a generalized model on the data from many clients, without fine-tuning for a target time series.

\section*{Approach} \label{sec:approach}

We use an encoder-decoder Transformer \cite{transformer} as a load forecasting model.
\diff{This model architecture has self-attention and cross-attention as its main . and was initially used for machine translation.
It was used as a forecasting model in \cite{time-series-transformer-prevalence} and later adopted for load forecasting \cite{time-augmented-transformer, ci-paper, tccml-paper}. We use the model implementation from \cite{ci-paper}.}

\begin{figure}[tb]
    \centering
    \includegraphics[width=0.85\textwidth]{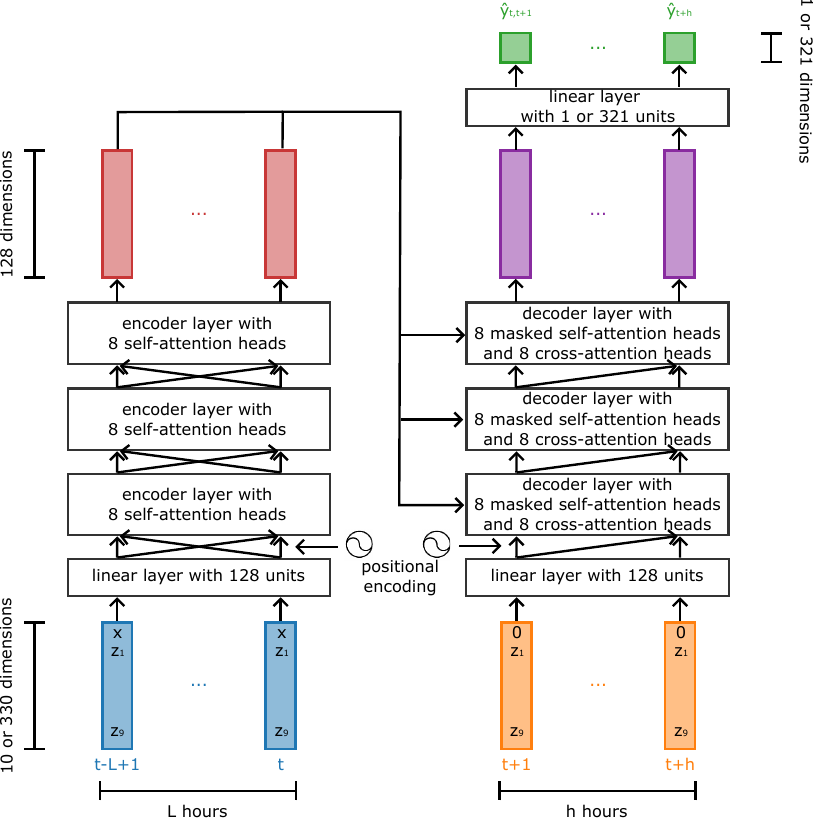}
    \caption{Architecture of the Transformer forecasting model. The input and output dimensions differ for the \mv\ model and the \loc\ and \glob\ models. \diff{The shown dimensions refer to the Electricity dataset with 321 clients.}}
    \label{fig:architecture}
\end{figure}

The encoder gets $L$ vectors as input, which represent the last $L$ time steps, where $L$ is called the lookback size.
Each input vector consists of one (in the case of \loc\ and \glob\ models) or $C$ (in the case of \mv\ models) load values, and nine additional time and calendar features.
The features are the hour of the day, the day of the week and the month (all cyclically encoded with a sine and a cosine function), whether it is a workday, whether it is a holiday and whether the next day is a workday (all binary features).
The input to the decoder consists of $h$ vectors, which represent the following $h$ time steps for which a forecast will be made.
In the decoder input, the load values are set to zero, so that each value is forecasted independently from the previous forecasted values, allowing for a direct multi-step forecast instead of generating all values iteratively.
The input vectors to the encoder and the decoder are first fed through linear layers to increase the dimensionality to the hidden dimension of the model $d_\mathrm{model}$.
Both the encoder and the decoder consist of multiple layers with eight self-attention heads and the decoder layers have eight additional masked cross-attention heads.
Finally, a linear layer transforms the $h$ decoder output vectors into a forecast with $h \times 1$ (for \loc\ and \glob\ models) or $h \times C$ (for \mv\ models) values.
We varied the number of encoder and decoder layers and the hidden dimension $d_\mathrm{model}$, and found three layers with $d_\mathrm{model} = 128$ to give the best results.
The full model architecture is shown in Figure \ref{fig:architecture}.

\subsection*{Training strategies} \label{sec:training-strategies}

We compare \mv, \loc\ and \glob\ Transformers.
The training strategies are depicted in Figure \ref{fig:training-strategies} and are further explained in the following.
Details on the inputs, outputs, number of models and training data size for each training strategy are given in Table \ref{tab:training-strategies}.

\begin{itemize}
    \item \emph{\Mv\ training strategy:} In the input to the model, each time step is represented by a vector of size $C + f$, where $C$ is the number of load time series and $f$ is the number of calendar features.
    The model forecasts $C$ values for the next $h$ time steps, i.e. its output consists of $h$ vectors of size $C$.
    A single model is used to forecast all time series simultaneously.
    \item \emph{\Loc\ training strategy:} \Loc\ models get only one time series as input and generate a forecast for this time series.
    In the input, each time step is represented by a vector with $f + 1$ entries for the $f$ calendar features and the electrical load value.
    $C$ separate models are trained for the $C$ time series, each using the training data from one time series.
    \item \emph{\Glob\ training strategy:} The \glob\ approach is a single model that generalizes for all load time series.
    The model gets one load time series as input and generates a forecast for that load time series.
    In contrast to the \loc\ models, only one \glob\ model is trained on samples from all load time series, and this model is used to forecast all load time series.
    This results in $C$ times as many training data for the \glob\ model as for a \loc\ model.
    To generate forecasts for all $C$ time series, the \glob\ model is used $C$ times with the history of one load time series as input.
\end{itemize}

\begin{table}
    \caption{Training strategy details for the \textit{Electricity} dataset with 321 load time series, 2.1 years training data and nine time and calendar features. For the \loc\ models, \textit{training data} is the amount of training data per model.}
    \label{tab:training-strategies}
    \centering
    \begin{tabular}{l|c|c|c|c}
        \hline
         Training strategy & models & input size & output size & training data \\
        \hline
         \Mv   &   1 & $L \times 330$ & $h \times 321$ & 2.1 years \\
         \Loc  & 321 & $L \times 10$ & $h \times 1$   & 2.1 years \\
         \Glob &   1 & $L \times 10$ & $h \times 1$   & 321 $*$ 2.1 years \\
        \hline
    \end{tabular}
\end{table}

\section*{Experimental setup} \label{sec:experiments}


\subsection*{Datasets} \label{sec:dataset}

\diff{As recommended in recent literature reviews on load forecasting \cite{haben-review-low-voltage-load-forecasting, hong-literature-review-outlook, vom-scheidt-data-analytics-review}, we conduct experiments on multiple datasets, namely the \textit{Electricity} and the \textit{Ausgrid solar home} datasets.
For both datasets we make a temporal split and use the first 70\% of each time series for training, the next 10\% for validation, and the last 20\% for testing, as in related work \cite{autoformer, fedformer, patchtst, transformers-effective}.}

The \textit{Electricity} dataset\footnote{\url{https://github.com/laiguokun/multivariate-time-series-data}} \diff{is published in \cite{long-term-datasets} and used in related work on long-term forecasting \citelongterm.}
It is a subset of the UCI Electricity Load Diagrams dataset\footnote{\url{https://archive.ics.uci.edu/ml/datasets/ElectricityLoadDiagrams20112014}} first presented in \cite{uci-dataset}, only containing the time series without missing values.
The dataset contains hourly electrical load data from 321 clients of a \diff{Portuguese} energy supplier.
The clients are from different economic sectors, including offices, factories, supermarkets, hotels, restaurants, among others \cite{uci-dataset}.
The time series range from 2012 to 2014.

\diff{The \textit{Ausgrid solar home} dataset\footnote{\url{https://www.ausgrid.com.au/Industry/Our-Research/Data-to-share/Solar-home-electricity-data}} contains solar generation and electrical load data from 300 clients\footnote{We use 299 of the clients because one client had missing data.} of an Australian energy supplier. The clients are private houses with rooftop solar systems. The time series range from July 2010 to June 2013. We only use the electrical load data transformed into hourly resolution.}

\subsection*{Comparison methods}

We compare our models with models from related work \citelongterm, as well as with a persistence baseline, linear regression models, multi-layer perceptrons and long short-term memory networks.
\begin{itemize}
    \item \emph{Models from related work:} For Informer \cite{informer}, Autoformer \cite{autoformer}, FEDformer \cite{fedformer}, PatchTST \cite{patchtst} and LTSF-Linear \cite{transformers-effective}, we take the results reported in the publications where applicable, and run the code published with the papers otherwise.
    All parameters except for the forecast horizon are left unchanged.
    \item \emph{Persistence baseline:} The persistence baseline takes the value from one week before the predicted hour as a forecast for the 24 hours and 96 hours horizons, and the value from one month before the predicted hour as the 720 hours forecast.
    \item \emph{Linear regression:} For each load time series, we train a linear regression model with $h$ outputs.
    The input consists of the last 336 load values and the nine time and calendar features for the current hour when the prediction is made (see \nameref{sec:approach} for a description of the features).
    The main difference to LTSF-Linear \cite{transformers-effective} is that the linear regression models are \loc\ models, but LTSF-Linear is a \glob\ model.
    Furthermore, the two approaches use different training algorithms and LTSF-Linear does not use time and calendar features.
    \item \emph{Multi-layer perceptron (MLP):} As for the linear regression, we train a \loc\ MLP for each load time series.
    The MLPs get the last 168 load values and the nine time and calendar features of the current hour as input.
    Using more than 168 load values as input did not improve the results.
    Each MLP has two hidden layers with ReLU activation \cite{relu} and 1024 neurons per layer.
    \item \diff{\emph{Long short-term memory (LSTM):} We train \mv, \loc\ and \glob\ LSTM \cite{lstm} models. We use the same architecture as in \cite{residential-load-forecasting-lstm}, consisting of two LSTM layers with 20 units each and a linear prediction layer. Using larger models did not improve the results.}
\end{itemize}

\subsection*{Training details}

All models are trained with the AdamW optimizer \cite{adamw} using the mean squared error loss.
We use a batch size of 128 and a learning rate of 0.0001 with 1000 warm-up steps and cosine decay with $\gamma = 0.8$.
When testing different lookback sizes $L$, we find one week to be optimal for the \mv\ Transformer and the \loc\ Transformers.
For the \glob\ Transformer, the results improve with increasing lookback size until $L=336$ (two weeks), and stay almost the same for $L=720$ (one month).
For Transformer models with two weeks input and one month output, the batch size has to be reduced to 64 due to the quadratic memory consumption of the model.
For the \mv\ Transformer, the batch size is set to 32 as in related work \cite{informer, autoformer, fedformer}.
The validation error is evaluated every 10,000 training steps and at the end of every epoch.
We use early stopping to end the training when no more improvement on the validation set is seen for ten evaluations.
For the MLPs, the initial learning rate is set to 0.001 and decayed with $\gamma = 0.5$ after every epoch.

\subsection*{Metric}

As in related work \cite{informer, autoformer, fedformer, patchtst, transformers-effective}, every load time series is standardized by subtracting its mean and dividing by its standard deviation and the metrics are computed on these standardized time series.
For every hour $t \in T_\mathrm{test}$ in the test set, a forecasting model predicts the next $h$ hourly loads $\hat{y}_t^c = \hat{y}_{t,t+1}^c, ..., \hat{y}_{t,t+h}^c$ for time series $c$.
Then, the mean absolute error (MAE) between the predictions $\hat{y}^c = \{\hat{y}_i^c\ \forall\ i \in T_\mathrm{test}$\} and the ground truth $y^c = y_1^c, ..., y_{T_\mathrm{test}}^c$ is computed.
As the final result, the MAE averaged across all $C$ load time series, the $T_\mathrm{test}$ evaluation time points and the $h$ forecasting steps is reported.
$$\mathrm{MAE(y, \hat{y}}) = \dfrac{1}{C \cdot |T_\mathrm{test}| \cdot h} \sum_{c=1}^C \sum_{t\in T_\mathrm{test}} \sum_{i=1}^{h} |y_{t+i}^c - \hat{y}_{t,t+i}^c|$$
\diff{The mean squared error (MSE) is computed analogously, using the squared residuals instead of the absolute residuals.}

\section*{Results} \label{sec:results}


\subsection*{Forecast accuracy}
\diff{Table \ref{tab:results} shows the MAE results on the two datasets\footnote{The MSE results show a similar pattern and can be found in appendix \ref{apx:additional-results}.}.
On the Electricity dataset, the global Transformer is the best model for the 24 hours horizon, and PatchTST is the best model for longer horizons.
On the Ausgrid solar home dataset, PatchTST is the best model for all three horizons.
The \glob\ Transformer beats the \loc\ Transformers and the \mv\ Transformer across all tested horizons. On average, it reduces the error by 21.8\% compared to the \mv\ Transformer and by 12.8\% compared to the \loc\ Transformers. Compared to the best local model, the linear regression, it reduces the error by 2.9\%. Compared to the best multivariate model, FEDformer, it reduces the error by 15.4\%.
All \mv\ models, including Informer, Autoformer, FEDformer and the \mv\ Transformer, perform poorly and do not beat the persistence baseline with a lag of one week.
The local linear regression models are slightly better than the global linear model, LTSF-Linear, on the Electricity dataset, but it is vice versa on the Ausgrid solar home dataset.
The MLP is in five out of six cases a bit worse than the linear regression, with a 1.5\% larger error on average.
The local LSTMs are better than the local Transformers, but the Transformer is better as a multivariate model and as a global model (except for the one month horizon on the Electricity dataset).
The forecast errors are lower on the Electricity dataset than on the Ausgrid dataset which is a more fine-grained dataset containing single private houses.}



\subsection*{Computational cost}
The training times are given in Table \ref{tab:results:runtime}.
The \loc\ Transformer models need by far the longest time to train.
Their training time increases sharply with longer forecast horizons.
The \mv\ Transformer trains fast and is even faster than the MLPs for short horizons.
Training a \glob\ Transformer is much faster than training the many \loc\ Transformers but takes longer than the linear regression, MLP and the \mv\ Transformer.
\diff{The LSTM trains always faster than the Transformer with the same training strategy.}

\newcommand{\bb}[1]{\textbf{#1}}
\newcommand{\ii}[1]{\textit{#1}}

\begin{table}[htb]
    \caption{MAE results on the two datasets, with 24, 96 and 720 hours forecast horizon. MV = \mv, L = \loc, G = \glob. The best results are highlighted in bold and the best results per training strategy are highlighted in italic.}
    \label{tab:results}
    \begin{adjustbox}{center}
    \difffigure{
    \begin{tabular}{lcc|ccc|ccc}
     \hline
     Model & strat- & input & \multicolumn{3}{c|}{Electricity} & \multicolumn{3}{c}{Ausgrid} \\
      & egy & (days) & 24h & 96h & 720h & 24h & 96h & 720h \\
     \hline
     Informer \cite{informer}                  & MV   & 4    & 0.399 & 0.407 & 0.450 & 0.582 & 0.607 & 0.645 \\
     Autoformer \cite{autoformer}              & MV   & 4    & 0.289 & 0.317 & 0.361 & 0.579 & 0.569 & \ii{0.592} \\
     FEDformer \cite{fedformer}                & MV   & 4    & \ii{0.284} & \ii{0.297} & \ii{0.343} & \ii{0.560} & \ii{0.566} & 0.609 \\
     LSTM                                      & MV   & 7    & 0.400 & 0.402 & 0.407 & 0.611 & 0.618 & 0.613 \\
     Transformer                               & MV   & 7    & 0.366 & 0.384 & 0.382 & 0.584 & 0.586 & 0.576\\
     \hdashline
     Persistence                               & L    & -    & 0.279 & 0.279 & 0.447 & 0.647 & 0.647 & 0.717 \\
     Linear regression                         & L    & 14   & 0.203 & \ii{0.233} & \ii{0.296} & \ii{0.496} & \ii{0.524} & \ii{0.565} \\
     MLP                                       & L    & 7    & \ii{0.199} & 0.236 & 0.308 & 0.499 & 0.532 & 0.567 \\
     LSTM                                      & L    & 7    & 0.263 & 0.283 & 0.337 & 0.517 & 0.541 & 0.573 \\
     Transformer                               & L    & 7    & 0.256 & 0.289 & 0.354 & 0.535 & 0.563 & 0.583 \\
     \hdashline
     LTSF-Linear \cite{transformers-effective} & G    & 14   & 0.209 & 0.237 & 0.301 & 0.490 & 0.515 & 0.553 \\
     PatchTST \cite{patchtst}                  & G    & 14   & 0.190 & \bb{0.222} & \bb{0.290} & \bb{0.468} & \bb{0.494} & \bb{0.522} \\
     LSTM                                      & G    & 7    & 0.207 & 0.239 & 0.302 & 0.491 & 0.525 & 0.559  \\
     Transformer                               & G    & 14   & \bb{0.184} & 0.225 & 0.312 & 0.482 & 0.514 & 0.533 \\
     \hline
    \end{tabular}
    }
    \end{adjustbox}
\end{table}

\begin{table}[bt]
    \caption{Training times in hours, measured on a machine with a Nvidia 3090 RTX GPU.}
    \label{tab:results:runtime}
    \difffigure{
    \begin{adjustbox}{center}
    \begin{tabular}{l|rrr|rrr}
        \hline
        Model & \multicolumn{3}{c|}{Electricity} & \multicolumn{3}{c}{Ausgrid}\\ 
         & 24h & 96h & 720h & 24h & 96h & 720h \\
        \hline
        Linear regression (\loc) &  0.02 &  0.03 &   0.08 & 0.02 & 0.03 &  0.07 \\
        MLP (\loc)               &  0.42 &  0.42 &   0.36 & 0.40 & 0.39 &  0.39 \\
        LSTM (\mv)               &  0.06 &  0.08 &   0.30 & 0.03 & 0.04 &  0.10 \\
        LSTM (\loc)              &  8.25 &  7.61 &   7.20 & 4.27 & 3.55 &  3.49 \\
        LSTM (\glob)             &  1.09 &  0.82 &   0.98 & 1.11 & 0.84 &  0.71 \\
        Transformer (\mv)        &  0.19 &  0.23 &   0.88 & 0.10 & 0.09 &  0.39 \\
        Transformer (\loc)       & 14.20 & 16.82 & 102.09 & 8.33 & 9.74 & 62.53 \\
        Transformer (\glob)      &  3.42 &  2.00 &   9.86 & 3.85 & 2.85 &  6.77 \\
        \hline
    \end{tabular}
    \end{adjustbox}
    }
\end{table}

\section*{Discussion} \label{sec:discussion}








\diff{
\paragraph{Best Transformer training strategy:}
On the two datasets, the \glob\ Transformer is superior to the \mv\ and \loc\ Transformers.
We hypothesize that this is a result of the larger number of training samples for the \glob\ model (see Table \ref{tab:training-strategies}).
The Transformer benefits from more training data, even if the training data comes from different sources.
The \mv\ models on the other hand are prone to overfitting.}

\diff{
\paragraph{Best Transformer architecture:}
PatchTST is the best model in five out of six cases. However, the difference to the \glob\ Transformer is small. This shows that the success of PatchTST is mainly a result of its \glob\ training strategy. Its improvement upon the \glob\ Transformer can be due to the patching mechanism, a better hyperparameter configuration, or the encoder-only architecture.
Among the \mv\ models, Autoformer \cite{autoformer} and FEDformer \cite{fedformer} give better results than the \mv\ Transformer.
It remains an open question whether these architectures are also better \glob\ models than the standard Transformer and PatchTST \cite{patchtst}.
Another promising architecture is the Temporal Fusion Transformer \cite{temporal-fusion-transformer}.
In previous work with just one aggregated time series, the Informer \cite{informer} also gave better results than the Transformer \cite{ci-paper}.
}

\paragraph{Comparison with the state of the art:}
The \glob\ Transformer achieves a better result for short-term forecasting on the Electricity dataset than related work \citelongterm, and achieves close results to the best results from PatchTST \cite{patchtst} for longer horizons \diff{and on the Ausgrid solar home dataset}.
However, to establish a state of the art for short-term and medium-term load forecasting, a comparison to other forecasting models must be undertaken, including models that are not based on the Transformer architecture and that are more sophisticated than our baselines.
Using weather data could improve the forecasts, because some electrical load patterns, such as the usage of electrical heating, are weather-dependent.
Weather features could affect which model gives the best results, because some models might be better in capturing these dependencies than others.



\paragraph{Linear models:}
As in related work \cite{transformers-effective}, we observe that linear models are strong baselines.
The linear regression is in five out of six cases the best \loc\ model and only outperformed by the \loc\ MLP for the one day horizon on the Electricity dataset.
No general answer can be given on whether the local linear regression models are better or the global LTSF-Linear is better, because each variant is better on one dataset.

\paragraph{Task complexity:}
For longer horizons, the \glob\ Transformer's performance compared to the linear models deteriorates. This can be due to the increasing complexity when the model forecasts many values simultaneously.
We chose a direct multi-step forecasting model because good results were achieved with this procedure before \cite{patchtst, transformers-effective}. However, other multi-step forecasting procedures, such as iterative single-step and iterative multi-step forecasting \cite{comparison-multi-step-strategies, comparative-analysis-multi-step}, could be beneficial for long-term forecasting because they reduce the number of forecasted values per model run. 

\paragraph{Transfer learning:}
According to the definition of transfer learning in \cite{transfer-learning-smart-buildings-review}, the \glob\ training strategy can be seen as a transfer learning method, because the model must transfer knowledge between different types of buildings with different consumption patterns.
Pre-training on other tasks than forecasting or on less similar data from domains other than electricity, as well as fine-tuning for a time series of interest, could improve the results.
An advantage of the \glob\ model is that it can be applied to new time series without retraining.
In \cite{tccml-paper} it was shown that the Transformer generalizes better to new time series than other approaches, but the forecasts are still better when training data from the target time series is available.

\paragraph{Other forecasting tasks:}
The Transformer model and the different training strategies are not designed for load forecasting in particular, but can also be applied to other forecasting tasks. We hypothesize that the \glob\ training strategy can also be beneficial for other datasets containing multiple time series with similar patterns.

\section*{Conclusion and future work} \label{sec:conclusion}

 We compare three Transformer training strategies for load forecasting on \diff{two datasets with multiple years of data for multiple hundred clients.}
 We show that the \mv\ training strategy used in related work on forecasting with Transformers \cite{informer, autoformer, fedformer} is not optimal, and it is better to use a \glob\ model instead.
 This shows that the right training strategy is crucial to get good results from a Transformer.
 Our approach achieves better results than related work \cite{informer, autoformer, fedformer}, and comes close to the best results from PatchTST \cite{patchtst}.
 In particular, our approach gives better results than the linear models from \cite{transformers-effective} for one day to four days forecasting horizons, which shows that, with the right training strategy, Transformers are effective for load forecasting.
 However, simple linear models give decent results for both short-term and medium-term horizons and train much faster than the Transformers.
 
 In the future, more sophisticated Transformer architectures could be tested with the \glob\ training strategy.
 A comparison to other forecasting methods could be undertaken, and weather data could be incorporated into the models to see how it affects the results.
 Experiments with other datasets and varying amounts of training data could show under which circumstances the \glob\ Transformer model is better than other approaches.
 Additionally, transfer learning from other tasks and datasets could be tested.
Future work could experiment with different datasets with varying amounts of data
to see how much training data is needed for the \glob\ model to surpass the \loc\ models.
A compromise between local and global models could be established by first clustering similar time series and then training one global model per cluster. The cluster-specific models would have less training data than the \glob\ model, but could benefit from the training data being more similar.
Potentially, the \glob\ training strategy could also be beneficial \diff{for other forecasting tasks than load forecasting.}

\section*{Acknowledgement}

This project is funded by the Helmholtz Association’s Initiative and Networking Fund through Helmholtz AI, the Helmholtz Association under the Program “Energy System Design”, and the German Research Foundation (DFG) as part of the Research Training Group 2153 “Energy Status Data: Informatics Methods for its Collection, Analysis and Exploitation”.

\section*{Availability of data and materials} 
  See section \nameref{sec:dataset} for the sources of the public datasets.
  Code is available on GitHub via \url{https://github.com/KIT-IAI/transformer-training-strategies}.

\section*{Author's contributions}
  MH: Conceptualisation, Investigation, Methodology, Software, Validation, Visualisation, Writing - original draft.
  MB: Writing - original draft, Writing - review and editing.
  BH, ON: Writing - review and editing.
  BS, RM, VH: Funding acquisition, Supervision, Writing - review and editing.

\section*{Competing interests}
  The authors declare that they have no competing interests.

\bibliographystyle{ieeetr}
\bibliography{exampleEI23}

\appendix

\section{Additional results} \label{apx:additional-results}

The mean squared errors for the same models as in Table \ref{tab:results} are given in Table \ref{tab:additional-results}.

\begin{table}[htb]
    \caption{MSE results on the two datasets, with 24, 96 and 720 hours forecast horizon. The best results are highlighted in bold and the best results per training strategy are highlighted in italic.}
    \label{tab:additional-results}
    \begin{adjustbox}{center}
    \difffigure{
    \begin{tabular}{lcc|ccc|ccc}
     \hline
     Model & strat- & input & \multicolumn{3}{c|}{Electricity} & \multicolumn{3}{c}{Ausgrid} \\
      & egy & (days) & 24h & 96h & 720h & 24h & 96h & 720h \\
     \hline
     Informer \cite{informer}                  & MV   & 4    & 0.305 & 0.320 & 0.384 & 0.853 & 0.870 & 0.891 \\
     Autoformer \cite{autoformer}              & MV   & 4    & 0.166 & 0.201 & 0.254 & 0.732 & 0.754 & 0.800 \\
     FEDformer \cite{fedformer}                & MV   & 4    & \ii{0.164} & \ii{0.183} & \ii{0.231} & \ii{0.707} & \ii{0.741} & \ii{0.798} \\
     LSTM                                      & MV   & 7    & 0.311 & 0.314 & 0.329 & 0.841 & 0.851 & 0.835 \\
     Transformer                               & MV   & 7    & 0.267 & 0.292 & 0.300 & 0.853 & 0.841 & 0.800\\
     \hdashline
     Persistence                               & L    & -    & 0.214 & 0.214 & 0.490 & 1.163 & 1.163 & 1.353 \\
     Linear regression                         & L    & 14   & 0.103 & \ii{0.133} & \bb{0.194} & \ii{0.604} & \ii{0.659} & \ii{0.725} \\
     MLP                                       & L    & 7    & \ii{0.097} & 0.135 & 0.212 & 0.627 & 0.693 & 0.749 \\
     LSTM                                      & L    & 7    & 0.146 & 0.169 & 0.234 & 0.647 & 0.693 & 0.751 \\
     Transformer                               & L    & 7    & 0.144 & 0.186 & 0.271 & 0.721 & 0.776 & 0.802 \\
     \hdashline
     LTSF-Linear \cite{transformers-effective} & G    & 14   & 0.110 & 0.140 & 0.203 & 0.598 & 0.647 & 0.705 \\
     PatchTST \cite{patchtst}                  & G    & 14   & 0.094 & 0.129 & \ii{0.197} & \bb{0.576} & \bb{0.641} & \bb{0.704} \\
     LSTM                                      & G    & 7    & 0.106 & 0.139 & 0.202 & 0.603 & 0.667 & 0.719  \\
     Transformer                               & G    & 14   & \bb{0.090} & \bb{0.127} & 0.219 & 0.599 & 0.665 & 0.716 \\
     \hline
    \end{tabular}
    }
    \end{adjustbox}
\end{table}

\end{document}